\documentclass[runningheads]{llncs}

\usepackage{eccv}

\usepackage{eccvabbrv}

\usepackage{graphicx}
\usepackage{booktabs}
\usepackage{multirow}
\usepackage{subcaption}
\usepackage{bm}     
\usepackage{amsmath}
\usepackage{amssymb}
\usepackage{pifont}
\usepackage{xcolor}
\usepackage{tikz}
\usetikzlibrary{calc}
\usepackage{placeins}
\usepackage{enumitem} 
\newcommand{\cmark}{\textcolor{green!70!black}{\ding{51}}}
\newcommand{\xmark}{\textcolor{red}{\ding{55}}}

\newcommand{\ours}{SpectralSplat}

\usepackage{hyperref}

\begin{document}

\title{\ours{}: Appearance-Disentangled Feed-Forward Gaussian Splatting\\for Driving Scenes} 

\titlerunning{\ours}

\author{Quentin Herau\inst{1}$^{*}$ \and
Tianshuo Xu\inst{1,2}$^{*\dagger}$ \and
Depu Meng\inst{1}$^{\S}$ \and
Jiezhi Yang\inst{1} \and
Chensheng Peng\inst{1,3}$^{\dagger}$ \and
Spencer Sherk\inst{1} \and
Yihan Hu\inst{1} \and
Wei Zhan\inst{1,3}$^{\ddagger}$}

\authorrunning{Q.~Herau et al.}

\institute{Applied Intuition \and
The Hong Kong University of Science and Technology \and
University of California, Berkeley\\[6pt]
{\small $^{*}$Equal contribution. $^{\S}$Project lead. $^{\dagger}$Work done as an intern at Applied Intuition. $^{\ddagger}$Corresponding author: \texttt{wei.zhan@applied.co}}}

\maketitle

\begin{abstract} \label{sec:abstract}
Feed-forward 3D Gaussian Splatting methods have achieved impressive reconstruction quality for autonomous driving scenes, yet they entangle scene geometry with transient appearance properties such as lighting, weather, and time of day.
This coupling prevents relighting, appearance transfer, and consistent rendering across multi-traversal data captured under varying environmental conditions.
We present \ours{}, a method that disentangles appearance from geometry within a feed-forward Gaussian Splatting framework.
Our key insight is to factor color prediction into an appearance-agnostic \emph{base} stream and an appearance-conditioned \emph{adapted} stream, both produced by a shared MLP conditioned on a global appearance embedding derived from DINOv2 features.
To enforce disentanglement, we train with paired observations generated by a hybrid relighting pipeline that combines physics-based intrinsic decomposition with diffusion-based generative refinement, and supervise with complementary consistency, reconstruction, cross-appearance, and base color losses.
We further introduce an appearance-adaptable temporal history that stores appearance-agnostic features, enabling accumulated Gaussians to be re-rendered under arbitrary target appearances.
Experiments demonstrate that \ours{} preserves the reconstruction quality of the underlying backbone while enabling controllable appearance transfer and temporally consistent relighting across driving sequences.
\end{abstract}
\section{Introduction}
\label{sec:intro}

Reconstructing 3D driving scenes from multi-camera video is a cornerstone capability for autonomous vehicle simulation, planning validation, and closed-loop testing~\cite{yang2023unisim}.
Recent feed-forward 3D Gaussian Splatting (3DGS) methods~\cite{chen2024mvsplat,charatan2024pixelsplat,shi2025unisplat} have made remarkable progress: a single forward pass through a learned network produces a complete set of 3D Gaussian primitives, enabling real-time novel view synthesis without costly per-scene optimization.
Among these, UniSplat~\cite{shi2025unisplat} achieves state-of-the-art quality on large-scale driving benchmarks by fusing multi-view spatial and multi-frame temporal information through a unified 3D latent scaffold.

However, current feed-forward methods fundamentally \emph{entangle} scene appearance with geometry.
The predicted Gaussian colors are ``baked in'', tightly coupled to the specific lighting, weather, and exposure conditions present in the input images.
This entanglement creates three practical limitations.
First, it precludes appearance editing: one cannot relight a reconstructed scene or transfer the appearance of a sunset to a daytime capture.
Second, it undermines temporal accumulation: when a streaming history buffer aggregates Gaussians from past frames that were observed under slightly different illumination, the result exhibits inconsistent coloring across the reconstructed scene.
Third, it prevents effective use of multi-traversal driving data, where the same road segment is captured repeatedly under diverse environmental conditions, a rich data source that could improve reconstruction coverage and robustness.

\input{figures/teaser}

We introduce \ours{}, a method that separates appearance from geometry within a feed-forward 3DGS framework for driving scenes (Fig.~\ref{fig:teaser}).
The core idea draws on the analogy of spectral decomposition: just as a spectrum reveals the individual wavelengths hidden within white light, \ours{} decomposes a scene representation into appearance-agnostic geometry and a controllable appearance code that can be freely recombined.

Concretely, we augment a feed-forward Gaussian Splatting backbone~\cite{shi2025unisplat} with three components.
(1)~A \emph{global appearance embedding} computed from DINOv2~\cite{oquab2024dinov2} patch tokens captures scene-level appearance characteristics (lighting mood, color cast, weather tone) in a compact latent vector shared across all Gaussians.
(2)~A \emph{factored color prediction} scheme evaluates a shared color MLP twice: once with a zeroed embedding to produce a canonical \emph{base color} (appearance-agnostic), and once with the actual embedding to produce an \emph{adapted color} matching the input conditions.
(3)~An \emph{appearance-adaptable temporal history} stores appearance-agnostic features alongside cached Gaussian geometry, so that accumulated primitives from previous timesteps can be re-rendered under the current target appearance rather than being locked to stale colors.

Training requires paired data showing the same scene under different appearances.
We obtain these pairs through a hybrid relighting pipeline that fuses physics-based intrinsic priors~\cite{wu2025mvinverse} with diffusion-based refinement~\cite{iclight}, and supervise with four complementary losses that jointly enforce clean separation of appearance from geometry.

Table~\ref{tab:method_comparison} positions \ours{} among existing approaches.
Our contributions are as follows:
\begin{itemize}[label=\textbullet]
    \item We propose a global appearance embedding with factored color prediction, enabling controllable appearance transfer without modifying the geometry pathway.
    \item We design a paired supervision framework using hybrid-relighted image pairs to enforce clean disentanglement of appearance from geometry.
    \item We introduce a temporal history mechanism that stores appearance-agnostic features, enabling accumulated Gaussians to be coherently re-rendered under any target appearance.
    \item We develop a hybrid relighting pipeline combining physics-based intrinsic priors with diffusion-based refinement via frequency-aware latent guidance, producing multi-view consistent training pairs.
\end{itemize}

\begin{table}[t]
\centering
\small
\caption{\textbf{Method comparison.} We compare along three axes: \emph{Feed-forward}~-- inference without per-scene optimization; \emph{Disentangled}~-- appearance is disentangled from geometry; \emph{Novel}~-- can render under unseen target appearances (\eg from a reference image). \ours{} is the only method satisfying all three.}
\label{tab:method_comparison}
\setlength{\tabcolsep}{4pt}
\vspace{-2mm}
\begin{tabular}{l ccc}
\toprule
Method & Feed-forward & Disentangled & Novel \\
\midrule
NeRF~\cite{mildenhall2020nerf}, 3DGS~\cite{kerbl2023gaussian}  & \xmark & \xmark & \xmark \\
NeRF-W~\cite{martinbrualla2021nerfw}, WildGaussians~\cite{kulhanek2024wildgaussians}, SWAG~\cite{dahmani2024swag} & \xmark & \cmark & \xmark \\
StyleRF~\cite{liu2023stylerf}                  & \xmark & \cmark & \cmark \\
MVSplat~\cite{chen2024mvsplat}, DepthSplat~\cite{xu2025depthsplat}, UniSplat~\cite{shi2025unisplat} & \cmark & \xmark & \xmark \\
\midrule
\textbf{\ours{} (ours)}                        & \cmark & \cmark & \cmark \\
\bottomrule
\end{tabular}
\vspace{-2mm}
\end{table}

\section{Related Work}
\label{sec:related}

\textbf{Optimization-based and Feed-forward 3D Reconstruction.}
Traditional pipelines for 3D reconstruction and novel view synthesis typically follow two stages: camera parameter estimation via Structure-from-Motion~\cite{schonberger2016colmap,snavely2006photo} and stereo matching~\cite{furukawa2009furu,schonberger2016pixelwise}, followed by differentiable rendering via NeRF~\cite{mildenhall2020nerf} and its variants~\cite{barron2022mip,kerbl2023gaussian,barron2023zip,wang2021neus,huang20242d,herau2023moisst,herau2024soac,herau20243dgs,herau2025pose, yang2024carff} to optimize neural scene representations per scene.
While accurate, these methods require minutes to hours of optimization per scene, limiting their scalability.
Recent feed-forward methods bypass per-scene optimization by directly regressing 3D representations from visual inputs.
Approaches such as Splatter Image~\cite{szymanowicz2024splatter}, PixelSplat~\cite{charatan2024pixelsplat}, MVSplat~\cite{chen2024mvsplat}, and DrivingRecon~\cite{lu2024drivingrecon} predict Gaussian primitives from posed images in a single forward pass.
This paradigm has been extended to pose-free settings by NoPoSplat~\cite{ye2024no} and Splatt3R~\cite{smart2024splatt3rzeroshotgaussiansplatting}, to causal streaming by Spann3R~\cite{wang20253d} and PreF3R~\cite{chen2024pref3rposefreefeedforward3d}, and to joint structure-and-pose inference by VGGT~\cite{wang2025vggt} and AnySplat~\cite{jiang2025anysplat}.
For driving scenes specifically, UniSplat~\cite{shi2025unisplat} achieves state-of-the-art quality by fusing multi-view and multi-frame information through a 3D latent scaffold with a dual-branch Gaussian decoder.
While highly scalable, all these feed-forward architectures produce appearance-entangled reconstructions: the predicted Gaussian colors are locked to the input lighting conditions, precluding relighting or appearance transfer.

\textbf{Appearance-Conditioned View Synthesis.}
To handle photometric variations across image collections, prior works incorporate per-image appearance embeddings to condition neural representations.
Block-NeRF~\cite{tancik2022block} and NeRF in the Wild~\cite{martinbrualla2021nerfw} learn latent appearance codes that account for global illumination and exposure changes.
WildGaussians~\cite{kulhanek2024wildgaussians} and SWAG~\cite{dahmani2024swag} extend 3D Gaussian Splatting to in-the-wild captures by augmenting Gaussians with trainable appearance embeddings and affine color transformations.
However, a critical limitation restricts these methods' applicability to autonomous driving: they fundamentally rely on dense multi-view coverage of the \textit{same} static scene captured under \textit{significantly different} lighting conditions to disentangle geometry from appearance.
Autonomous driving datasets consist of long, forward-facing trajectories where the ego-vehicle rarely revisits the exact same location.
This lack of multi-illumination supervision for identical viewpoints makes it difficult for embedding-based optimization methods to learn a disentangled lighting representation, often leading to overfitting or ``baked-in'' shadows.
Furthermore, these methods require per-scene optimization, forfeiting the scalability advantages of feed-forward inference.

\textbf{Generative Relighting and 3D Stylization.}
Early data-driven relighting approaches relied on GANs~\cite{isola2017image,zhu2017unpaired}, but operate strictly in 2D image space without 3D consistency.
Specialized portrait methods~\cite{sun2019single,pandey2021total} introduced explicit lighting representations, yet extending these to complex, unbounded scenes remains challenging.
Recent work has shifted toward physically grounded diffusion models.
In the 2D domain, IC-Light~\cite{iclight} and Uni-Renderer~\cite{chen2025uni} impose light transport consistency to generate high-quality single-view relighting edits.
For multi-view consistency, 3D inverse rendering methods such as MVInverse~\cite{wu2025mvinverse} decompose scenes into intrinsic components (albedo, normals, roughness) for physics-based re-rendering.
Parallel to inverse rendering, 3D stylization approaches such as ARF~\cite{zhang2022arf} and StyleRF~\cite{liu2023stylerf} lift 2D style transformations into 3D feature fields to ensure coherence across viewpoints.
However, 2D diffusion models lack inherent geometric consistency required for multi-view video synthesis, while physics-based methods often produce high-frequency artifacts due to the ill-posed nature of inverse rendering, particularly in unbounded outdoor scenes.
\ours{} addresses this gap by combining a feed-forward 3D Gaussian backbone with explicit appearance conditioning, enabling consistent relighting on dynamic, single-pass driving sequences without per-scene optimization.

\section{Method}
\label{sec:method}

We present \ours{}, a method that disentangles appearance from geometry within a feed-forward Gaussian Splatting framework.
Our approach modifies the color prediction pathway of the UniSplat~\cite{shi2025unisplat} backbone while keeping the geometry inference.

% We first describe the backbone (Sec.~\ref{sec:ff3d}), then our three architectural contributions: a global appearance embedding (Sec.~\ref{sec:global_latent}), factored color prediction (Sec.~\ref{sec:two_stage_color}), and appearance-adaptable temporal history (Sec.~\ref{sec:history}).
% We then explain the hybrid relighting pipeline that generates paired training data (Sec.~\ref{sec:relighting}) and the paired supervision framework that enforces disentanglement (Sec.~\ref{sec:paired_supervision}).

\subsection{Feed-Forward Gaussian Splatting Backbone}
\label{sec:ff3d}

Our method builds on UniSplat~\cite{shi2025unisplat}, a feed-forward 3DGS framework for driving scenes. We briefly summarize the components relevant to our contributions; full architectural details are in~\cite{shi2025unisplat} and the supplementary.

UniSplat uses a frozen Pi3~\cite{wang2025pi3permutationequivariantvisualgeometry} geometry transformer for dense 3D point prediction and a DINOv2-S~\cite{oquab2024dinov2} backbone for semantic features.
These are fused via a feature pyramid network (FPN) to produce per-pixel image features $\mathbf{z}_i^{\mathrm{img}} \in \mathbb{R}^{D_{\mathrm{img}}}$.
The predicted points are voxelized and processed by a sparse 3D U-Net with temporal history fusion, yielding per-voxel features $\mathbf{z}_j^{\mathrm{vox}} \in \mathbb{R}^{D_{\mathrm{vox}}}$.
While Pi3 remains frozen, DINOv2, the FPN, and the 3D U-Net are fine-tuned end-to-end.

A dual-branch decoder then predicts 3D Gaussians.
The \emph{voxel branch} decodes $K$ Gaussians per voxel from $\mathbf{z}_j^{\mathrm{vox}}$ alone.
The \emph{point branch} decodes one Gaussian per pixel from the concatenation of $\mathbf{z}_i^{\mathrm{img}}$ and a voxel feature sampled from the fused scaffold at the corresponding 3D position, giving each point Gaussian access to both view-specific and 3D context.
Both branches predict geometry (position offset, scale, rotation, opacity, dynamic score).

We do not modify the geometry pathway: all geometric parameters are produced as in UniSplat.
Color prediction, however, is factored into appearance-agnostic and appearance-aware components as described below.

\begin{figure}[t]
    \centering
    \includegraphics[width=0.95\textwidth]{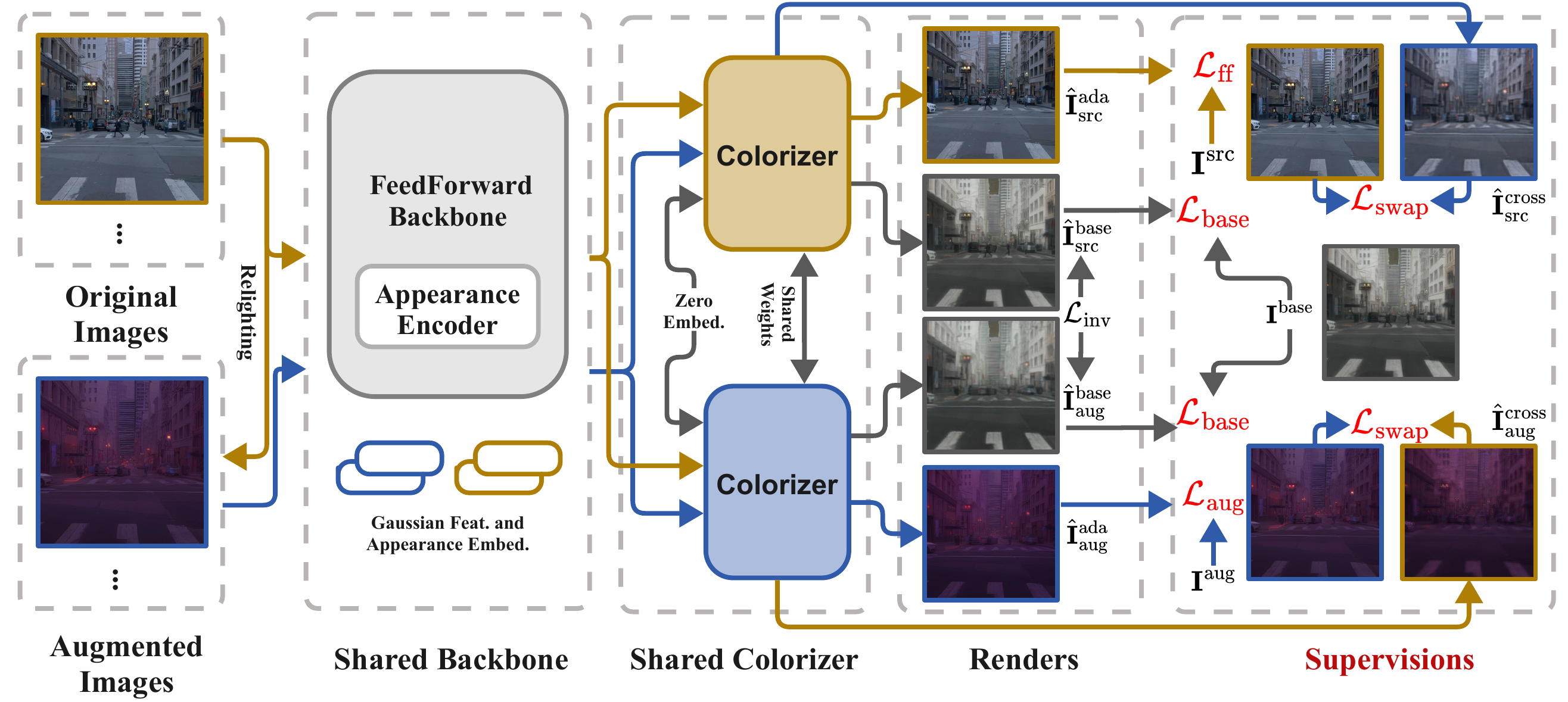}
    \vspace{-2mm}
    \caption{\textbf{Training pipeline.} Original and augmented images share geometry but produce separate features and appearance embeddings. Four losses enforce disentanglement: $\mathcal{L}_{\mathrm{inv}}$ (base invariance), $\mathcal{L}_{\mathrm{aug}}$ (augmented reconstruction), $\mathcal{L}_{\mathrm{swap}}$ (cross-appearance), and $\mathcal{L}_{\mathrm{base}}$ (base color alignment).}
    \vspace{-2mm}
    \label{fig:training}
\end{figure}

\subsection{Global Appearance Embedding}
\label{sec:global_latent}

We introduce a global appearance embedding to capture scene-level appearance characteristics such as lighting conditions, weather, and time of day.
The embedding is derived from the DINOv2 features already extracted by the backbone.
Specifically, the projected $2048$-dimensional DINOv2 patch tokens for each camera view are globally average-pooled to obtain a single feature vector per camera, which is then passed through a lightweight encoder $\phi$:
\begin{equation}
    \mathbf{a}_k = \phi\!\left(\frac{1}{N_p}\sum_{n=1}^{N_p}\mathbf{f}_{k,n}\right),\quad
    \mathbf{a} = \frac{1}{|C_t|}\sum_{k\in C_t}\mathbf{a}_k \;\in\;\mathbb{R}^d,
\end{equation}
where $\mathbf{f}_{k,n} \in \mathbb{R}^{2048}$ are the projected DINOv2 patch tokens for camera~$k$, $C_t$ is the set of cameras at the current timestamp, and $\phi$ is a lightweight MLP encoder (architecture details in the supplementary).
We use $d{=}64$.
The resulting embedding $\mathbf{a}$ is shared across all Gaussians in the scene, enforcing a consistent color transformation across views.
Although UniSplat supervises on future frames ($t{+}1$), the appearance embedding is computed only from current-timestamp cameras.
Future-frame renders reuse the current embedding, which is the desired behavior: it forces the model to produce geometry that generalizes across viewpoints while the appearance code captures only the global illumination state at time~$t$. The disentanglement losses ($\mathcal{L}_{\mathrm{inv}}, \mathcal{L}_{\mathrm{swap}}, \mathcal{L}_{\mathrm{base}}$) are likewise applied only to current-frame renders, since augmented pairs are not generated for future frames.

This design reflects the observation that appearance changes in driving scenes (tone, color cast, illumination mood) are predominantly global.
A single shared latent imposes this prior directly and prevents per-point appearance drift.

\subsection{Factored Color Prediction}
\label{sec:two_stage_color}

Instead of predicting RGB directly from geometric features, we factor color prediction into two evaluations of a shared colorizer MLP.
We illustrate the mechanism on the \emph{point branch}; the voxel and sky branches follow the same structure, differing only in input features.

Each point Gaussian's feature is the concatenation of the voxel feature sampled from the fused scaffold ($\mathbf{z}_i^{\mathrm{vox}} \!\in\! \mathbb{R}^{D_{\mathrm{vox}}}$) and the per-pixel FPN image feature ($\mathbf{z}_i^{\mathrm{img}} \!\in\! \mathbb{R}^{D_{\mathrm{img}}}$).
The color MLP $f_{\mathrm{rgb}}^{\mathrm{pt}}$ is evaluated twice:
\begin{align}
    \mathbf{c}_i^{\mathrm{base}} &= \sigma\!\bigl(f_{\mathrm{rgb}}^{\mathrm{pt}}([\mathbf{z}_i^{\mathrm{vox}};\,\mathbf{z}_i^{\mathrm{img}};\,\mathbf{0}])\bigr), \label{eq:base_rgb}\\
    \mathbf{c}_i^{\mathrm{ada}} &= \sigma\!\bigl(f_{\mathrm{rgb}}^{\mathrm{pt}}([\mathbf{z}_i^{\mathrm{vox}};\,\mathbf{z}_i^{\mathrm{img}};\,\mathbf{a}])\bigr), \label{eq:ada_rgb}
\end{align}
where $\sigma$ is a sigmoid, $[\cdot;\cdot]$ concatenation, and $\mathbf{a}$ the global appearance embedding.
The \textbf{base color} $\mathbf{c}^{\mathrm{base}}$ (computed with a zero embedding) captures appearance-agnostic content; the \textbf{adapted color} $\mathbf{c}^{\mathrm{ada}}$ adapts to the target appearance.
The voxel branch uses $\mathbf{z}_j^{\mathrm{vox}}$ alone, while sky Gaussians use $\mathbf{z}_i^{\mathrm{img}}$ alone (no voxel context).
We use a zero vector instead of a learned canonical embedding so the base color has a fixed, optimization-independent reference point.
Rendering uses $\mathbf{c}^{\mathrm{ada}}$ as the final color; $\mathbf{c}^{\mathrm{base}}$ serves as a disentanglement signal (Sec.~\ref{sec:paired_supervision}).

\subsection{Appearance-Adaptable Temporal History}
\label{sec:history}

UniSplat's backbone accumulates Gaussian primitives from previous timesteps via a temporal history buffer to improve spatial coverage~\cite{shi2025unisplat}.
In the original design, cached Gaussians retain their original colors, which causes inconsistencies when past and present frames were captured under different lighting.
We modify this mechanism so that accumulated Gaussians can be \emph{re-rendered} under the current target appearance at recall time.

For the \emph{voxel branch}, temporal fusion already occurs at the feature level inside the 3D U-Net (historical voxel features are warped and fused with the current scaffold), so voxel RGB is always predicted fresh with the current appearance embedding---no modification is needed.

For the \emph{point branch}, we cache the per-pixel FPN image feature $\mathbf{z}_i^{\mathrm{img}}$ alongside each static point Gaussian's geometric parameters.
When recalled at a later timestep, the geometry is transformed to the current coordinate frame, fresh voxel context is obtained by querying the current fused scaffold, and the color is \emph{recomputed} with the current appearance embedding:
\begin{equation}
    \mathbf{c}_i^{\mathrm{hist}} = \sigma\!\bigl(f_{\mathrm{rgb}}^{\mathrm{pt}}([\mathbf{z}_i^{\mathrm{vox,fused}};\,\mathbf{z}_i^{\mathrm{img,cached}};\,\mathbf{a}^{\mathrm{now}}])\bigr).
\end{equation}
This enables newly predicted and historical Gaussians to be rendered under the same target appearance, eliminating the color inconsistency that would arise if accumulated Gaussians were locked to the appearance under which they were originally generated.

\subsection{Multi-View Consistent Relighting}
\label{sec:relighting}

Training \ours{} requires paired observations showing the same scene content under different appearances.
We generate such pairs using a hybrid relighting pipeline that synergizes physics-based intrinsic decomposition with diffusion-based generative priors.
Physics-based methods ensure multi-view 3D consistency via geometric constraints but often yield high-frequency artifacts due to the ill-posed nature of inverse rendering.
Conversely, diffusion models excel at photorealism (generating atmospheric effects and complex transport phenomena) but lack inherent 3D consistency across views.
We bridge these paradigms via a frequency-guided denoising strategy, which injects 3D-consistent low-frequency structural priors into the diffusion process while retaining the model's capacity for high-fidelity detail synthesis.

\noindent\textbf{Intrinsic Decomposition and Physics-Based Priors.}
Given multi-view images and camera poses, we employ MVInverse~\cite{wu2025mvinverse} to estimate per-pixel base color $\mathbf{A}_v$ and surface normals $\mathbf{N}_v$.
From these, we compute a Lambertian re-rendering $\hat{\mathbf{I}}_v^{\text{mv}}$ under a target lighting direction (derivation in supplementary).
This physics-based reference $\hat{\mathbf{I}}_v^{\text{mv}}$ guarantees global illumination consistency across views but lacks photorealistic high-frequency details (\eg specularities, sky textures), so we use it as a structural guidance signal for the generative stage.

\noindent\textbf{Generative Refinement via IC-Light.}
We refine $\hat{\mathbf{I}}_v^{\text{mv}}$ with IC-Light~\cite{iclight}, a relighting diffusion model adapted from Stable Diffusion~\cite{rombach2022high}. While IC-Light produces photorealistic lighting effects, applying it independently per view breaks multi-view consistency.

\noindent\textbf{Frequency-Aware Latent Guidance.}
To reconcile 3D consistency with perceptual quality, we intervene in the DDIM~\cite{song2020denoising} sampling trajectory via spectral decoupling.
We decompose any VAE latent $\mathbf{z}$ into low-frequency (structural) and high-frequency (textural) components using a Gaussian low-pass operator $\mathcal{G}_\sigma$: $\mathbf{z}_{\text{low}} {=} \mathcal{G}_\sigma(\mathbf{z})$, $\mathbf{z}_{\text{high}} {=} \mathbf{z} {-} \mathbf{z}_{\text{low}}$.
Let $\mathbf{z}_{\text{ref}}^{\text{low}} = \mathcal{G}_\sigma(\mathcal{E}(\hat{\mathbf{I}}_v^{\text{mv}}))$ be the low-frequency prior from the physics-based reference.
At each denoising step~$t$, we estimate the clean latent $\hat{\mathbf{z}}_0 = (\mathbf{z}_t - \sqrt{1-\bar{\alpha}_t}\,\boldsymbol{\epsilon}_\theta)/\sqrt{\bar{\alpha}_t}$ and replace only its low-frequency component with a guided version:
\begin{equation}
    \tilde{\mathbf{z}}_0 = \bigl[ (1 {-} w(t)) \cdot \mathcal{G}_\sigma(\hat{\mathbf{z}}_0) + w(t) \cdot \mathbf{z}_{\text{ref}}^{\text{low}} \bigr] + \bigl[ \hat{\mathbf{z}}_0 - \mathcal{G}_\sigma(\hat{\mathbf{z}}_0) \bigr],
    \label{eq:guided_x0}
\end{equation}
where $w(t) = \lambda \cdot (t/T)$ is a linearly decaying schedule ($\lambda \in [0,1]$).
Early denoising steps receive strong structural guidance that locks illumination direction and tone; later steps relax the constraint so the model can refine textures and atmospheric details.
The noise is then recomputed as $\tilde{\boldsymbol{\epsilon}} = (\mathbf{z}_t - \sqrt{\bar{\alpha}_t}\,\tilde{\mathbf{z}}_0)/\sqrt{1-\bar{\alpha}_t}$ to keep the trajectory on the DDIM deterministic manifold.
Operating in latent space, a small kernel ($\sigma {\approx} 2$) suffices to separate global illumination from local texture, ensuring multi-view geometric consistency while preserving the photorealism of the generative prior.

\begin{figure}[t]
    \centering
    \includegraphics[width=0.9\linewidth]{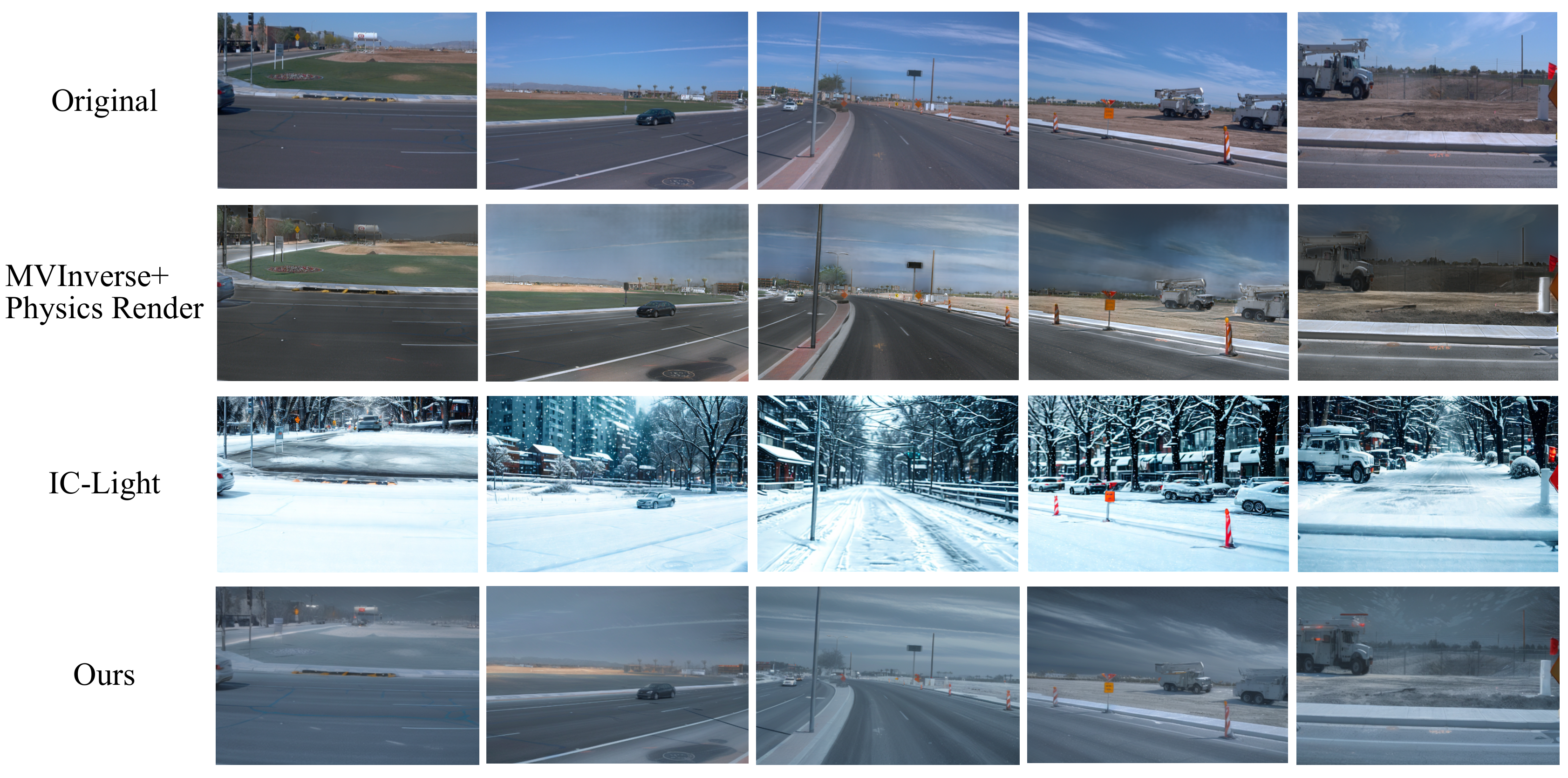}\\[2pt]
    \vspace{-2mm}
    \caption{\textbf{Relighting pipeline.} MVInverse~+~physics rendering is 3D-consistent but flat; IC-Light alone is photorealistic but inconsistent; our hybrid pipeline achieves both.}
    \vspace{-1mm}
    \label{fig:relight}
\end{figure}

Figure~\ref{fig:relight} compares our hybrid pipeline against its individual components and shows the diversity of relighting outputs that serve as paired training data for the supervision framework described next.

\subsection{Paired Appearance Supervision}
\label{sec:paired_supervision}

Using the augmented images $\mathbf{I}^{\mathrm{aug}}$ produced by the relighting pipeline (Sec.~\ref{sec:relighting}), we train with paired observations that share identical scene content but differ in appearance (Fig.~\ref{fig:training}).
Both original $\mathbf{I}^{\mathrm{src}}$ and augmented $\mathbf{I}^{\mathrm{aug}}$ images are processed by the shared backbone, yielding two distinct appearance embeddings $\mathbf{a}^{\mathrm{src}}$ and $\mathbf{a}^{\mathrm{aug}}$.
Geometry (point positions, voxel grid coordinates) is derived from the original depth predictions and shared between both pipelines; the FPN features, U-Net processing, and appearance embeddings are computed independently.

\subsection{Training Objective}
\label{sec:overall_objective}

All disentanglement losses are computed on non-sky regions.
Dynamic-region supervision (BCE on predicted vs.\ ground-truth masks) is applied to both original and augmented renders.

\noindent\textbf{Base Invariance.}
\label{sec:base_consistency}
Since base colors use a zero embedding, they should be appearance-invariant:
$\mathcal{L}_{\mathrm{inv}} = \|\hat{\mathbf{I}}^{\mathrm{base}}_{\mathrm{src}} - \hat{\mathbf{I}}^{\mathrm{base}}_{\mathrm{aug}}\|_2^2$.

\noindent\textbf{Augmented Reconstruction.}
The adapted render must reproduce the augmented target:
$\mathcal{L}_{\mathrm{aug}} = \|\hat{\mathbf{I}}^{\mathrm{ada}}_{\mathrm{aug}} - \mathbf{I}^{\mathrm{aug}}\|_2^2$.

\noindent\textbf{Appearance-Swap Consistency.}
\label{sec:cross_appearance}
At each step we randomly pick one of two cross-appearance directions (augmented features with original embedding, or vice versa) and supervise against the corresponding ground truth:
$\mathcal{L}_{\mathrm{swap}} = \|\hat{\mathbf{I}}^{\mathrm{cross}} - \mathbf{I}^{\mathrm{target}}\|_2^2$.
Only one direction is computed per step to halve memory, since both provide equivalent gradient signal in expectation.

\noindent\textbf{Base color Consistency.}
\label{sec:base_color_consistency}
We regularize base colors toward the pseudo-ground-truth base color from MVInverse:
$\mathcal{L}_{\mathrm{base}} = \|\hat{\mathbf{I}}^{\mathrm{base}}_{\mathrm{src}} - \mathbf{A}^{\mathrm{src}}\|_2^2$.

\noindent\textbf{Full objective.}
\begin{equation}
\begin{split}
    \mathcal{L} ={} &
    \underbrace{\lambda_m\,\mathcal{L}_{\mathrm{mse}} + \lambda_p\,\mathcal{L}_{\mathrm{lpips}} + \lambda_d\,\mathcal{L}_{\mathrm{dyn}} + \lambda_s\,\mathcal{L}_{\mathrm{depth}}}_{\text{Original feedforward supervision } \mathcal{L}_{\mathrm{ff}}} \\
    & + \beta_1\,\mathcal{L}_{\mathrm{inv}}
    + \beta_2\,\mathcal{L}_{\mathrm{aug}}
    + \beta_3\,\mathcal{L}_{\mathrm{swap}}
    + \beta_4\,\mathcal{L}_{\mathrm{base}},
\end{split}
\end{equation}
with $\lambda_m{=}5.0$, $\lambda_p{=}0.05$, $\lambda_d{=}0.05$, $\lambda_s{=}0.1$, $\beta_1{=}1.0$, $\beta_2{=}5.0$, $\beta_3{=}0.5$, $\beta_4{=}0.5$.
$\mathcal{L}_{\mathrm{mse}}$ and $\mathcal{L}_{\mathrm{lpips}}$ supervise the \emph{adapted} render only; the base stream receives gradients exclusively from $\mathcal{L}_{\mathrm{inv}}$ and $\mathcal{L}_{\mathrm{base}}$, preventing the photometric loss from pushing appearance information into the canonical color pathway.

\section{Experiments}
\label{sec:experiments}

\subsection{Experimental Setup}
\label{sec:exp_setup}

\noindent\textbf{Datasets.}
We evaluate \ours{} on two large-scale autonomous driving benchmarks.
\emph{Waymo Open Dataset}~\cite{sun2020scalability} comprises 798 training and 202 validation sequences with five surround-view cameras.
\emph{nuScenes}~\cite{caesar2020nuscenes} provides six surround-view cameras per frame; following~\cite{shi2025unisplat}, we partition scenes into equally spaced bins along the vehicle trajectory, yielding 135{,}941 training and 30{,}080 validation bins.
On Waymo, following UniSplat~\cite{shi2025unisplat}, we assess two tasks: \emph{reconstruction} (rendering input views at the current timestep~$t$) and \emph{novel view synthesis} (rendering at the subsequent timestep~$t{+}1$).
On nuScenes, we evaluate on target views consisting of the first, last, and central frames of each bin.

\noindent\textbf{Metrics.}
We report standard image quality metrics: PSNR, SSIM~\cite{wang2004image}, and LPIPS~\cite{zhang2018unreasonable}.
For cross-appearance evaluation, we additionally measure the reconstruction quality when rendering with either original or relighted image features and swapping the appearance embedding.

\noindent\textbf{Baselines.}
For reconstruction, we compare against feed-forward methods: UniSplat~\cite{shi2025unisplat}, MVSplat~\cite{chen2024mvsplat}, and DepthSplat~\cite{xu2025depthsplat} on Waymo; PixelSplat~\cite{charatan2024pixelsplat}, MVSplat, Omni-Scene~\cite{wei2025omni}, and UniSplat on nuScenes.
For appearance disentanglement, we compare against WildGaussians~\cite{kulhanek2024wildgaussians}, a per-scene optimization method with trainable appearance embeddings.

\noindent\textbf{Implementation Details.}
We adopt the same backbone architecture and training hyperparameters as UniSplat~\cite{shi2025unisplat}.
Unlike UniSplat, which trains the scale and shift predictor from scratch to align Pi3's affine-invariant depth to metric scale, we use LiDAR-derived ground-truth scale and shift for depth computation during training while still supervising the predictor with the same ground truth.
This decouples geometry quality from predictor convergence, enabling faster and more stable training; at inference, the learned predictor is used as usual without requiring LiDAR.
Augmented training pairs are generated offline using our hybrid relighting pipeline (Sec.~\ref{sec:relighting}).
All models are trained on 64 NVIDIA A100 GPUs with a batch size of 64 for 10 epochs.
\input{figures/qual_cross_appearance}

\subsection{Reconstruction Quality}
\label{sec:exp_recon}

We first verify that adding appearance disentanglement does not degrade reconstruction quality compared to the UniSplat baseline.
Table~\ref{tab:recon} reports standard metrics on both Waymo and nuScenes.
\ours{} outperforms UniSplat in most metrics.
These results demonstrate that factored color prediction does not degrade reconstruction quality.

\begin{table}[ht]
\caption{\textbf{Reconstruction quality.} \ours{} preserves reconstruction quality on both datasets while enabling appearance control. Baseline metrics reported from~\cite{shi2025unisplat}. Best in \textbf{bold}, second best \underline{underlined}.}
\vspace{-1mm}
\label{tab:recon}
\centering
\resizebox{0.85\linewidth}{!}{%
\scriptsize
\begin{tabular}{l ccc ccc c ccc}
\toprule
& \multicolumn{6}{c}{Waymo} & & \multicolumn{3}{c}{nuScenes} \\
\cmidrule(lr){2-7} \cmidrule(l){9-11}
& \multicolumn{3}{c}{Recon.} & \multicolumn{3}{c}{NVS} && \multicolumn{3}{c}{Recon. + NVS} \\
\cmidrule(lr){2-4} \cmidrule(lr){5-7} \cmidrule(l){9-11}
Method & PSNR & SSIM & LPIPS & PSNR & SSIM & LPIPS & & PSNR & SSIM & LPIPS \\
\midrule
PixelSplat~\cite{charatan2024pixelsplat} & --- & --- & --- & --- & --- & --- & & 21.51 & 0.616 & 0.372 \\
MVSplat~\cite{chen2024mvsplat}     & 24.94 & 0.80 & 0.23 & 22.04 & 0.68 & 0.34 & & 21.61 & 0.658 & 0.295 \\
DepthSplat~\cite{xu2025depthsplat} & 25.38 & 0.76 & 0.26 & 23.86 & 0.70 & 0.31 & & --- & --- & --- \\
Omni-Scene~\cite{wei2025omni}      & --- & --- & --- & --- & --- & --- & & 24.27 & 0.736 & \underline{0.237} \\
UniSplat~\cite{shi2025unisplat}    & \underline{29.58} & \underline{0.86} & \underline{0.17} & \textbf{25.98} & \textbf{0.76} & \underline{0.24} & & \underline{25.37} & \underline{0.765} & 0.246 \\
\midrule
\ours{}                             & \textbf{29.99} & \textbf{0.87} & \textbf{0.14} & \underline{25.83} & \textbf{0.76} & \textbf{0.23} & & \textbf{25.75} & \textbf{0.792} & \textbf{0.227} \\
\bottomrule
\end{tabular}%
}
\vspace{-1mm}
\end{table}

\subsection{Cross-Appearance Evaluation}
\label{sec:exp_cross_appearance}

We evaluate appearance disentanglement on the Waymo validation set.
For each sequence, we use the first 10 frames as input and generate augmented counterparts using our MVInverse~+~IC-Light relighting pipeline, providing paired original and augmented images with shared geometry but different appearances.

\noindent\textbf{Cross-appearance protocol.}
Given an original observation $\mathbf{I}^{\mathrm{src}}$ and its augmented counterpart $\mathbf{I}^{\mathrm{aug}}$, we render using four combinations of geometry features and appearance embeddings: two \emph{matched} (geometry and appearance from the same source) and two \emph{cross} (appearance embedding swapped to the other source).
Each render is compared against the ground truth matching its target appearance.
The gap between matched and cross PSNR directly quantifies how well appearance is separated from geometry.

\begin{table}[b]
\centering
\caption{\textbf{Cross-appearance evaluation.} Rows: geometry source; columns: appearance embedding. Diagonal entries (bold) are matched; off-diagonal entries measure cross-appearance transfer.}
\label{tab:cross_appearance}
\vspace{-2mm}
\setlength{\tabcolsep}{3pt}
\scriptsize
\begin{tabular}{ll ccc ccc}
\toprule
& & \multicolumn{3}{c}{Src appearance} & \multicolumn{3}{c}{Aug appearance} \\
\cmidrule(lr){3-5} \cmidrule(lr){6-8}
& & PSNR & SSIM & LPIPS & PSNR & SSIM & LPIPS \\
\midrule
\multirow{2}{*}{Waymo}
  & Src geom. & \textbf{29.99} & \textbf{0.875} & \textbf{0.143} & 23.42 & 0.728 & 0.439 \\
  & Aug geom. & 21.21 & 0.672 & 0.426 & \textbf{27.52} & \textbf{0.862} & \textbf{0.199} \\

\bottomrule
\end{tabular}
\vspace{-2mm}
\end{table}

The matched configurations (diagonal) achieve high reconstruction quality (29.99 and 27.52\,dB PSNR), confirming that the model faithfully reconstructs both original and augmented appearances.
The cross-appearance entries show that swapping the appearance embedding to a mismatched source still produces coherent renders (23.42 and 21.21\,dB), with the gap to matched quality reflecting the inherent difficulty of rendering under a different illumination.
Crucially, this swap is only possible because the appearance embedding captures global illumination independently of geometry---a capability that standard feed-forward methods like UniSplat lack entirely.

\begin{figure}[h]
    \centering
    \includegraphics[width=0.85\linewidth]{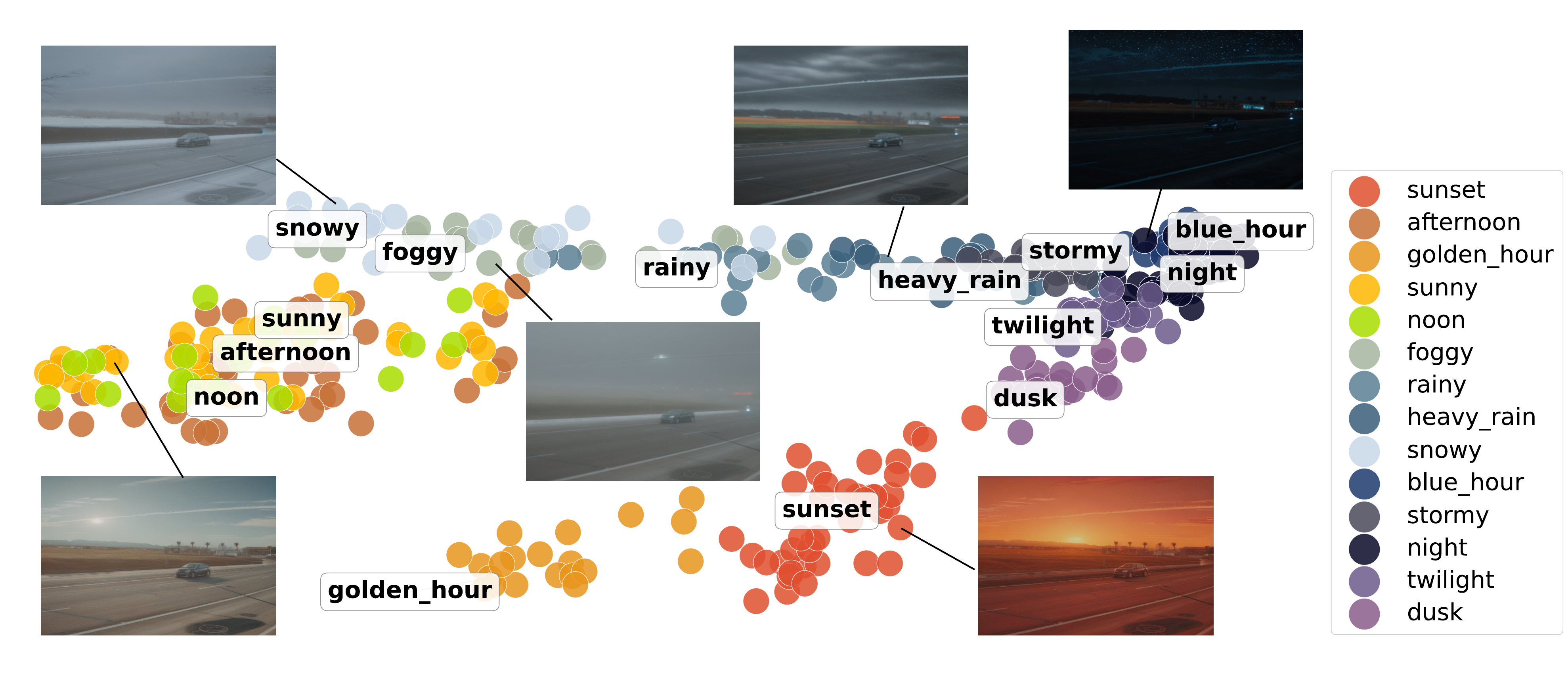}
    \caption{\textbf{t-SNE of appearance embeddings.} Embeddings cluster by illumination type, confirming the encoder captures meaningful appearance information.}
    \label{fig:tsne}
\end{figure}
\input{figures/qual_wildgaussians}
\begin{table}[h]
\centering
\caption{\textbf{Comparison with WildGaussians.} \ours{} achieves competitive appearance control in a feed-forward pass, ${\sim}$100$\times$ faster than per-scene optimization.}
\vspace{-2mm}
\label{tab:wild}
\resizebox{0.75\linewidth}{!}{%
\setlength{\tabcolsep}{4pt}
\begin{tabular}{l ccc c}
\toprule
Method & PSNR$\uparrow$ & SSIM$\uparrow$ & LPIPS$\downarrow$ & Time (per 10 frames) \\
\midrule
WildGaussians~\cite{kulhanek2024wildgaussians} & \textbf{30.42} & 0.843 & 0.296 & ${\sim}$31 min \\
\ours{} (ours)                                  & 27.52 & \textbf{0.862} & \textbf{0.199} & \textbf{${\sim}$18 s} \\
\bottomrule
\end{tabular}%
}
\vspace{-2mm}
\end{table}

\subsection{Comparison with Optimization-Based Methods}
\label{sec:exp_wild}

We compare \ours{} to WildGaussians~\cite{kulhanek2024wildgaussians}, an optimization-based method with trainable appearance embeddings.
We adapt WildGaussians to use a single per-frame embedding (shared across all cameras at a given timestep) instead of its default per-image embedding, and set the embedding dimension to $d{=}64$ to match our appearance code.
We train on the same Waymo sequences using the augmented images from the first 10 frames as the training set, and evaluate reconstruction quality on the augmented images.
While WildGaussians requires ${\sim}$31 minutes of optimization per scene, \ours{} performs inference on the whole scene in ${\sim}$18 seconds, making our method ${\sim}$100 times faster.

Figure~\ref{fig:qual_wg} compares \ours{} to WildGaussians on the same evaluation scenes.
Both methods are trained on augmented data, but WildGaussians tends to reproduce diffusion artifacts present in the IC-Light training images (halos, unnatural glow).
\ours{} produces perceptually cleaner renders with better geometric consistency. the disentanglement objective forces the model to learn the underlying appearance transformation rather than memorize noisy training targets.
This explains the metric profile in Table~\ref{tab:wild}: although \ours{} has lower PSNR, it achieves better SSIM and especially LPIPS, reflecting higher perceptual quality and structural fidelity.

\subsection{Qualitative Results}
\label{sec:exp_qualitative}

Figure~\ref{fig:tsne} shows a t-SNE visualization of the learned appearance embeddings, confirming that the latent space captures meaningful lighting variations.

Figure~\ref{fig:qual_cross} presents the full cross-appearance rendering on Waymo.
The adapted renders (rows~2,~5) closely match their respective ground truths, while the base color renders (rows~3,~6) are nearly identical despite originating from different inputs, visually confirming that $\mathcal{L}_{\mathrm{inv}}$ successfully enforces appearance-invariance.
The swap renders (rows~7,~8) transfer the target appearance to mismatched geometry, demonstrating that the embedding captures appearance independently.

\subsection{Ablation Study}
\label{sec:supp_ablation}

We ablate each component of \ours{} to quantify its contribution.
All ablations are trained for 2 epochs on Waymo (vs.\ 10 for the full model) and evaluated with the cross-appearance protocol (Sec.~\ref{sec:exp_cross_appearance}).

\begin{table}[t]
\centering
\caption{\textbf{Ablation study on Waymo (cross-appearance).} We report the swapped configurations (geometry from one source, appearance embedding from the other) to isolate each loss's contribution to disentanglement. Best in \textbf{bold}.}
\label{tab:ablation}
\vspace{-1mm}
\setlength{\tabcolsep}{4pt}
\scriptsize
\begin{tabular}{ll ccc}
\toprule
& & PSNR$\uparrow$ & SSIM$\uparrow$ & LPIPS$\downarrow$ \\
\midrule
\multirow{2}{*}{\ours{} (full)}
  & Src geom.\ $\to$ Aug app. & \textbf{21.84} & \textbf{0.678} & \textbf{0.502} \\
  & Aug geom.\ $\to$ Src app. & \textbf{19.16} & \textbf{0.579} & \textbf{0.514} \\
\midrule
\multirow{2}{*}{w/o $\mathcal{L}_{\mathrm{swap}}$}
  & Src geom.\ $\to$ Aug app. & 14.58 & 0.525 & 0.525 \\
  & Aug geom.\ $\to$ Src app. & 14.28 & 0.493 & 0.534 \\
\midrule
\multirow{2}{*}{w/o $\mathcal{L}_{\mathrm{base}}$}
  & Src geom.\ $\to$ Aug app. & 21.32 & 0.669 & 0.514 \\
  & Aug geom.\ $\to$ Src app. & 19.03 & 0.572 & 0.524 \\
\bottomrule
\end{tabular}
\vspace{-1mm}
\end{table}

Table~\ref{tab:ablation} reports cross-appearance metrics only (swapped geometry and appearance), directly measuring disentanglement quality.
The appearance-swap loss $\mathcal{L}_{\mathrm{swap}}$ is the most critical component: removing it causes cross-appearance PSNR to collapse from 21.84 to 14.58\,dB, indicating that the model memorizes each appearance independently rather than learning to separate it from geometry.
Removing the base color anchor $\mathcal{L}_{\mathrm{base}}$ degrades cross-appearance transfer, as the base color stream loses its grounding and the model shifts color information into the appearance embedding; this also prevents appearance-agnostic temporal accumulation (Sec.~\ref{sec:history}).
The full objective achieves the best cross-appearance transfer, confirming that the losses are complementary.

\section{Conclusion}
\label{sec:conclusion}

We presented \ours{}, a feed-forward 3DGS method that disentangles appearance from geometry for driving scenes.
A global appearance embedding conditions factored color MLPs to produce both appearance-agnostic and appearance-conditioned renders, while paired supervision from a hybrid relighting pipeline enforces clean separation without per-scene optimization.
Combined with an appearance-adaptable temporal history, \ours{} enables controllable appearance transfer on accumulated Gaussians in a single forward pass.

\noindent\textbf{Future Work.}
Promising extensions include spatially-varying appearance codes for finer-grained control (\eg per-object relighting, complex multi-source illumination), leveraging multi-traversal data where the same location is captured under naturally different conditions to reduce reliance on synthetic augmentation, and coupling appearance disentanglement with generative world models for fully controllable driving simulation.

\FloatBarrier

\clearpage
\section*{Supplementary Material}
\label{sec:supp}

\subsection*{Backbone Architecture Details}

We provide additional details on the UniSplat~\cite{shi2025unisplat} backbone that are relevant to understanding the feature dimensions used by our appearance modules.

\noindent\textbf{Feature Extraction.}
Pi3~\cite{wang2025pi3permutationequivariantvisualgeometry} predicts dense 3D point maps from multi-camera inputs.
DINOv2-S~\cite{oquab2024dinov2} extracts $384$-dimensional patch tokens, which are projected to a $2048$-dimensional space via a learned linear layer and combined with Pi3's intermediate representations through a multi-scale FPN, producing per-pixel image features of dimension $D_{\mathrm{img}}{=}256$.

\noindent\textbf{Voxel Branch.}
The predicted 3D points are voxelized onto a regular grid; each occupied voxel aggregates coordinates, input RGB, and projected FPN features.
A sparse 3D U-Net with encoder-decoder skip connections processes the voxels, incorporating temporal history fusion via ego-motion warping and sparse addition.
The output per-voxel features have dimension $D_{\mathrm{vox}}{=}32$, from which $K{=}2$ Gaussians per voxel are decoded by a geometry MLP.

\noindent\textbf{Point Branch.}
For each pixel, the per-pixel FPN feature ($D_{\mathrm{img}}{=}256$) is concatenated with a voxel feature ($D_{\mathrm{vox}}{=}32$) sampled from the nearest occupied voxel in the fused scaffold, yielding a $(D_{\mathrm{vox}}{+}D_{\mathrm{img}})$-dimensional feature vector.
A geometry MLP decodes position offset, scale, rotation quaternion, opacity, and dynamic score.

\subsection*{Appearance Module Details}

\noindent\textbf{Appearance Encoder.}
The appearance encoder~$\phi$ is a three-layer MLP ($2048 \!\to\! 256 \!\to\! 256 \!\to\! d$, $d{=}64$) with LayerNorm and GELU activations.
The projected $2048$-dimensional DINOv2 patch tokens for each camera are globally average-pooled before being passed through~$\phi$.
Per-camera embeddings are averaged over all cameras at the current timestamp to produce the global embedding~$\mathbf{a}$.

\noindent\textbf{Color MLPs.}
Each factored color MLP (point, voxel, sky) is a two-layer MLP with hidden dimension~128 and ReLU activations.
Input dimensions: $D_{\mathrm{vox}} + d = 96$ (voxel), $D_{\mathrm{vox}} + D_{\mathrm{img}} + d = 352$ (point), $D_{\mathrm{img}} + d = 320$ (sky).

\subsection*{Lambertian Relighting Derivation}

Given multi-view images $\{\mathbf{I}_v\}_{v=1}^{V}$ and camera poses $\{R_v\}_{v=1}^{V}$, MVInverse~\cite{wu2025mvinverse} estimates per-pixel base color $\mathbf{A}_v$ and normals $\mathbf{N}_v$.
To synthesize a target lighting environment defined by global direction $\boldsymbol{\ell}_w \in \mathbb{R}^3$, color $\mathbf{c} \in \mathbb{R}^3$, intensity $s$, and ambient term $k$, we compute Lambertian shading per view.
The light direction is transformed to camera coordinates via $\boldsymbol{\ell}_v = R_v \boldsymbol{\ell}_w$, yielding the shading map $\mathbf{S}_v = \max(0, \mathbf{N}_v \cdot \boldsymbol{\ell}_v)$.
The physics-based reference is:
\begin{equation}
    \hat{\mathbf{I}}_v^{\text{mv}} = \mathbf{A}_v \odot (s \cdot \mathbf{c} \odot \mathbf{S}_v + k),
\end{equation}
where $\odot$ denotes the Hadamard product.

% \subsection*{Implicit denoising of augmented data.}
% An interesting property of \ours{} is that it does not blindly reproduce the augmented training images.
% Because the model is supervised on \emph{both} original and augmented images through the shared factored color prediction, it learns the underlying appearance transformation---time of day, weather---rather than the diffusion artifacts introduced by IC-Light (\eg neon-like halos, unnatural glow).
% As illustrated in Fig.~\ref{fig:denoise_aug}, the adapted renders faithfully capture the target lighting mood while producing cleaner, more physically plausible images than the augmented ground truth itself.
% This implicit denoising effect is a direct consequence of the disentanglement objective: the base stream anchors geometry and texture, and the appearance embedding only needs to encode the global illumination shift.

% \begin{figure}[t]
%     \centering
%     \includegraphics[width=\linewidth,height=0.3\linewidth]{example-image-a}
%     \caption{\textbf{Implicit denoising.} Although trained on augmented images (middle) that contain diffusion artifacts, \ours{} renders (right) reproduce the target appearance without the noise, producing cleaner results than the training data. \textcolor{red}{[Placeholder.]}}
%     \label{fig:denoise_aug}
% \end{figure}

\subsection*{Appearance Transfer Grid}

We show in Figure~\ref{fig:style_grid} appearance transfer across four different scenes using five diverse appearance embeddings.
Each column applies the appearance embedding extracted from the reference image shown in the top row.
The first two columns show the original render and base color (zero embedding) for each scene.

\newlength{\gridcellw}
\setlength{\gridcellw}{0.14\textwidth}

\begin{figure*}[h]
\centering
\setlength{\tabcolsep}{0.5pt}
\renewcommand{\arraystretch}{0}
\begin{tabular}{@{}p{\gridcellw}p{\gridcellw}p{\gridcellw}p{\gridcellw}p{\gridcellw}p{\gridcellw}p{\gridcellw}@{}}
& &
\includegraphics[width=\gridcellw]{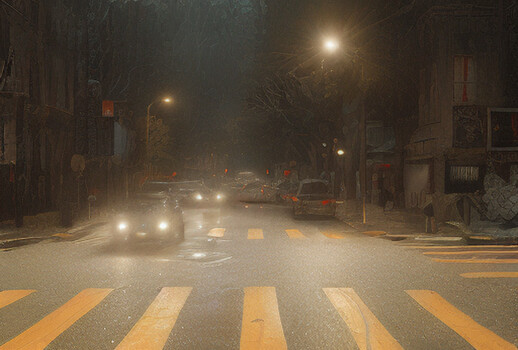} &
\includegraphics[width=\gridcellw]{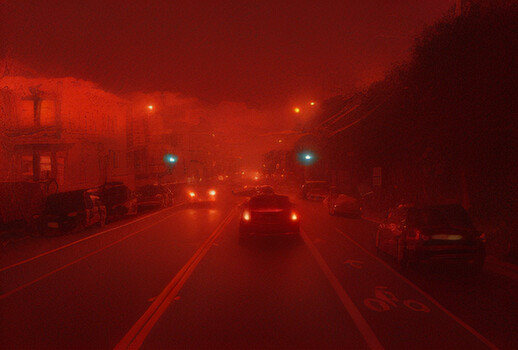} &
\includegraphics[width=\gridcellw]{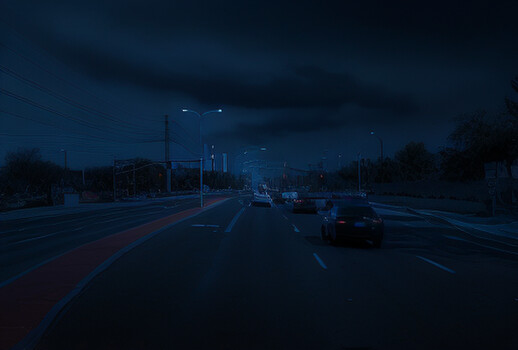} &
\includegraphics[width=\gridcellw]{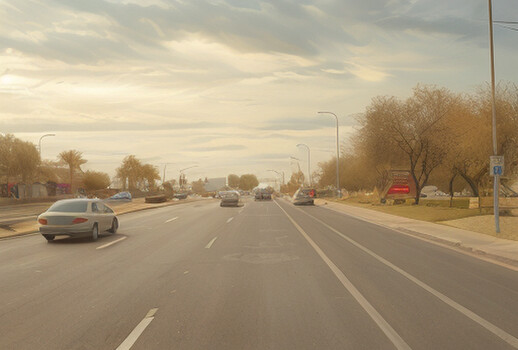} &
\includegraphics[width=\gridcellw]{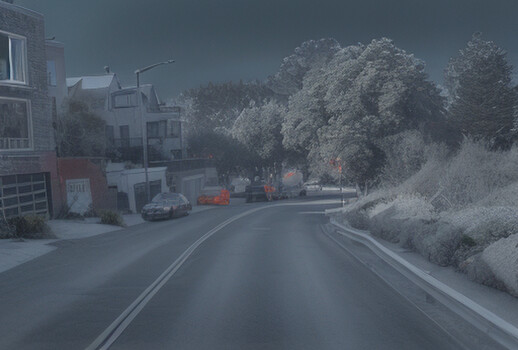} \\[0.5pt]
\includegraphics[width=\gridcellw]{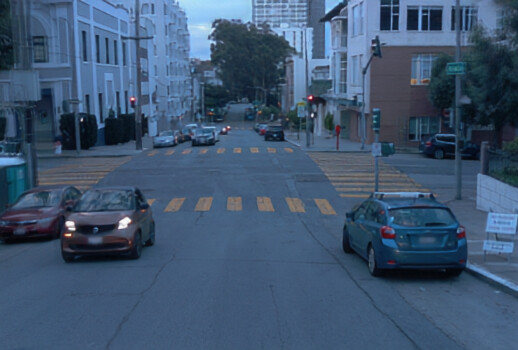} &
\includegraphics[width=\gridcellw]{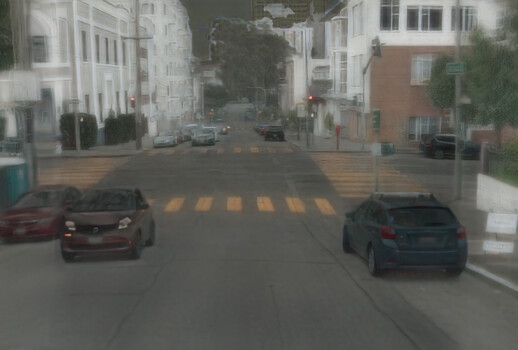} &
\includegraphics[width=\gridcellw]{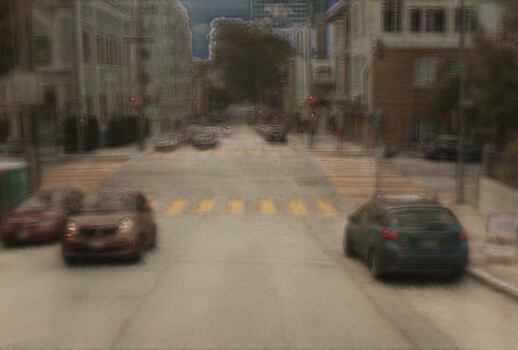} &
\includegraphics[width=\gridcellw]{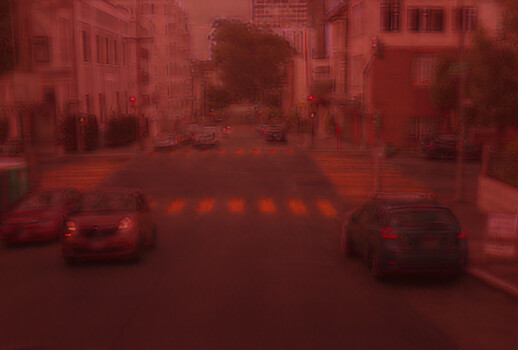} &
\includegraphics[width=\gridcellw]{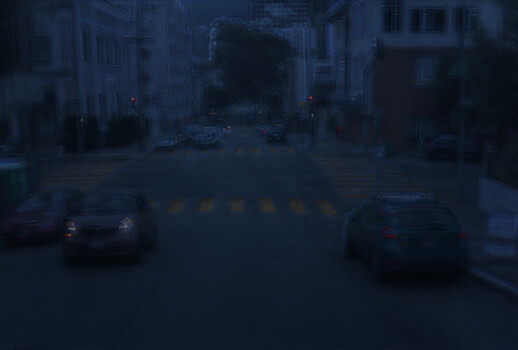} &
\includegraphics[width=\gridcellw]{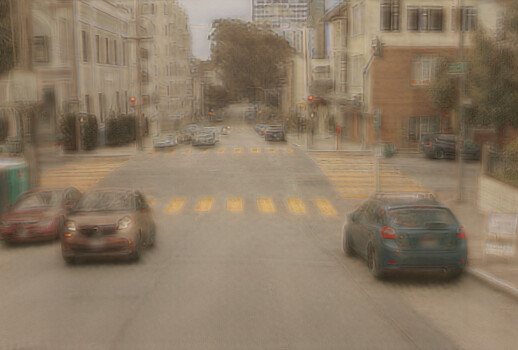} &
\includegraphics[width=\gridcellw]{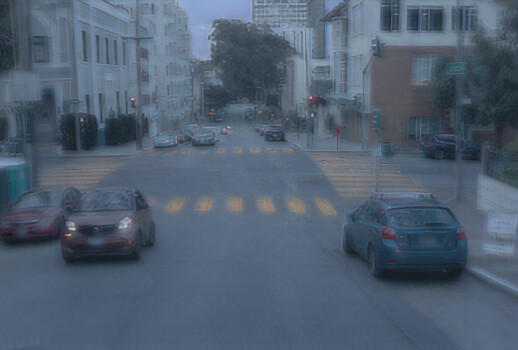} \\[0.5pt]
\includegraphics[width=\gridcellw]{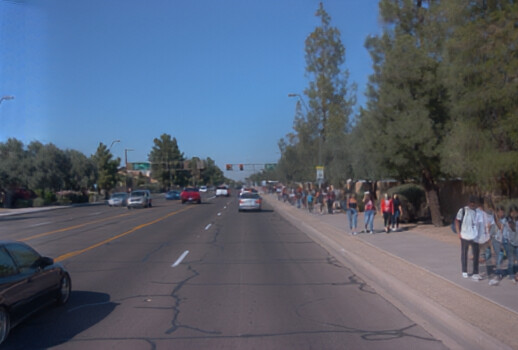} &
\includegraphics[width=\gridcellw]{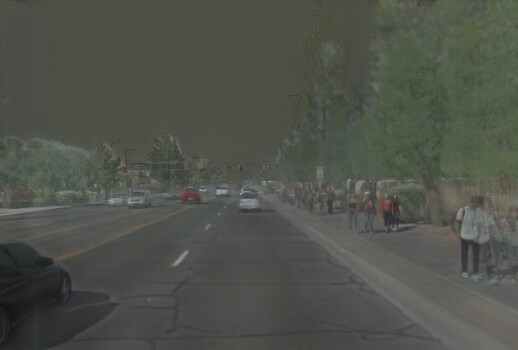} &
\includegraphics[width=\gridcellw]{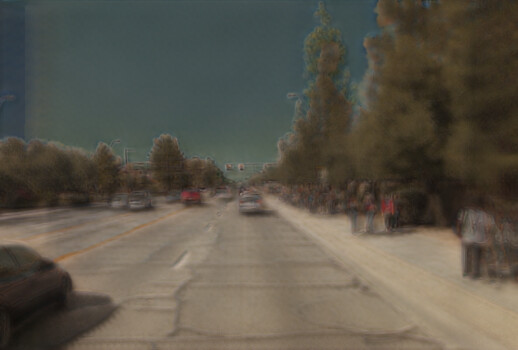} &
\includegraphics[width=\gridcellw]{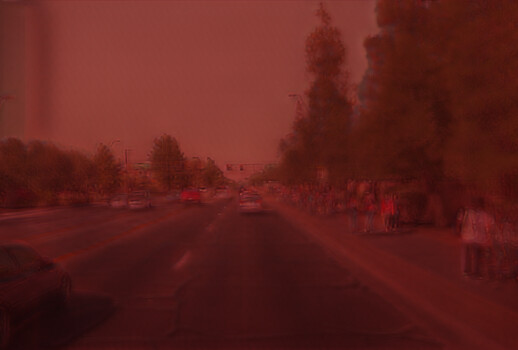} &
\includegraphics[width=\gridcellw]{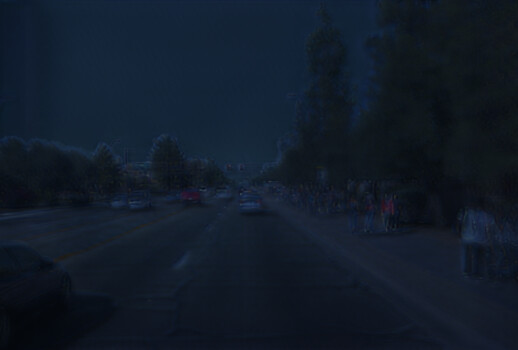} &
\includegraphics[width=\gridcellw]{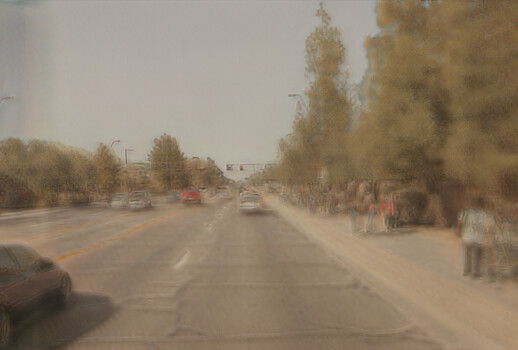} &
\includegraphics[width=\gridcellw]{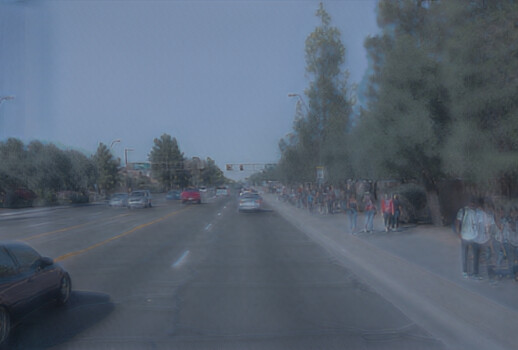} \\[0.5pt]
\includegraphics[width=\gridcellw]{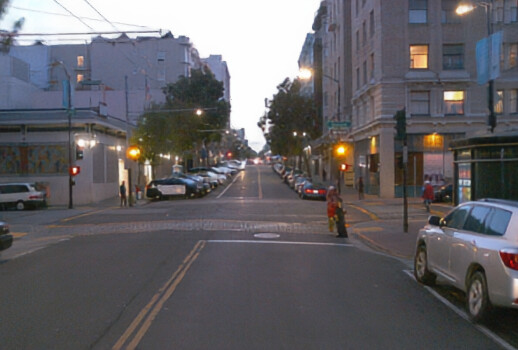} &
\includegraphics[width=\gridcellw]{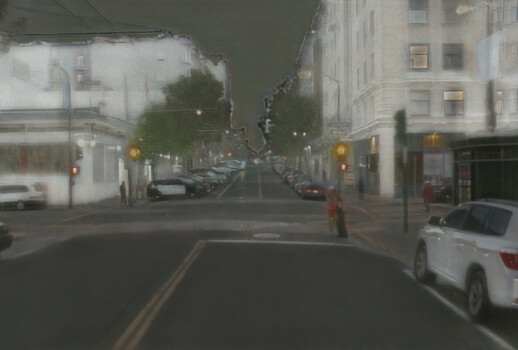} &
\includegraphics[width=\gridcellw]{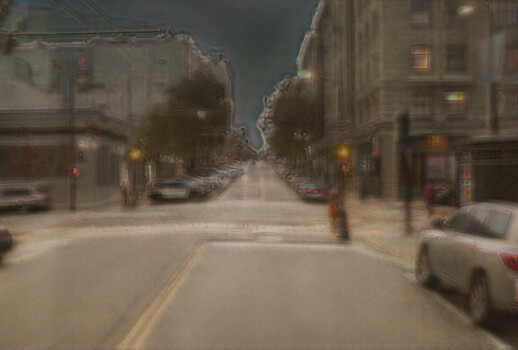} &
\includegraphics[width=\gridcellw]{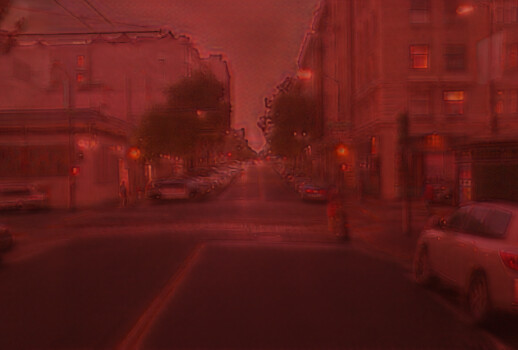} &
\includegraphics[width=\gridcellw]{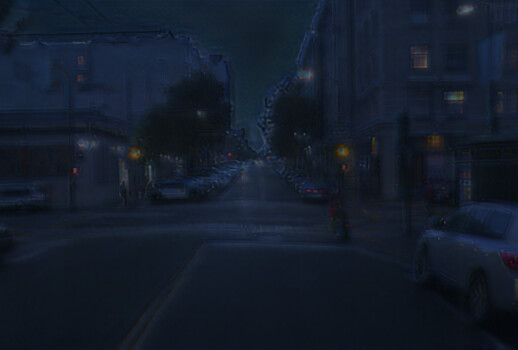} &
\includegraphics[width=\gridcellw]{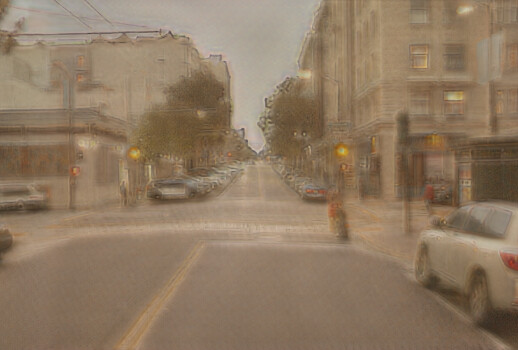} &
\includegraphics[width=\gridcellw]{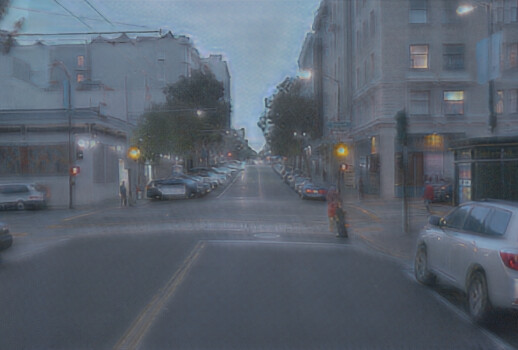} \\[0.5pt]
\includegraphics[width=\gridcellw]{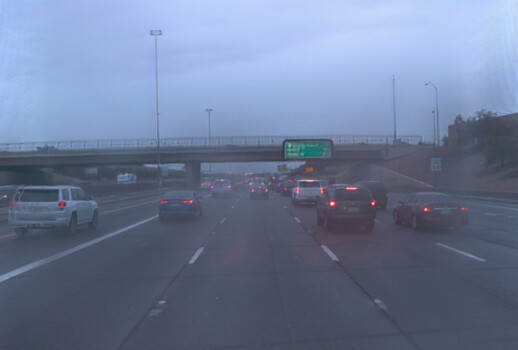} &
\includegraphics[width=\gridcellw]{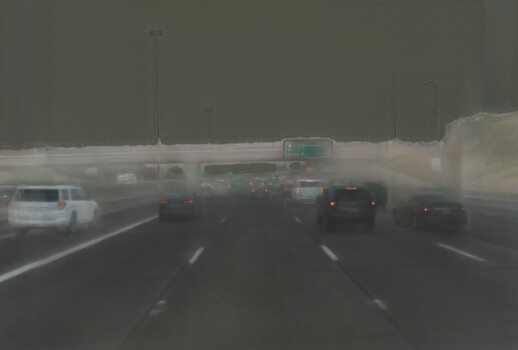} &
\includegraphics[width=\gridcellw]{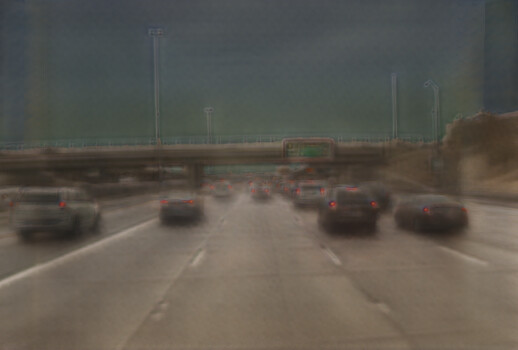} &
\includegraphics[width=\gridcellw]{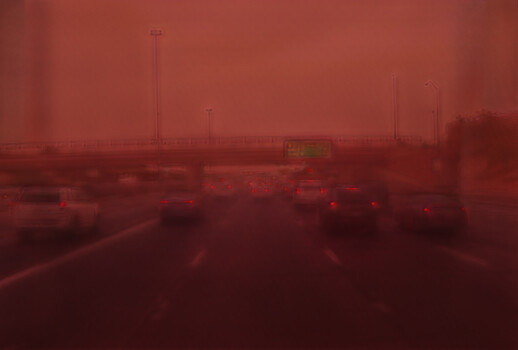} &
\includegraphics[width=\gridcellw]{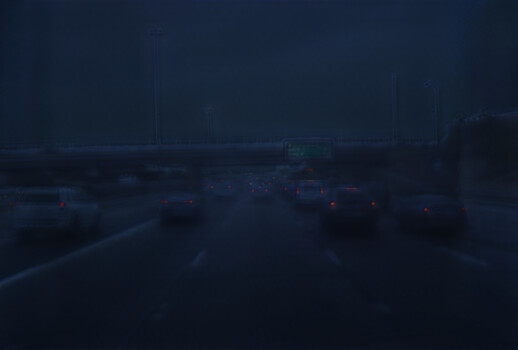} &
\includegraphics[width=\gridcellw]{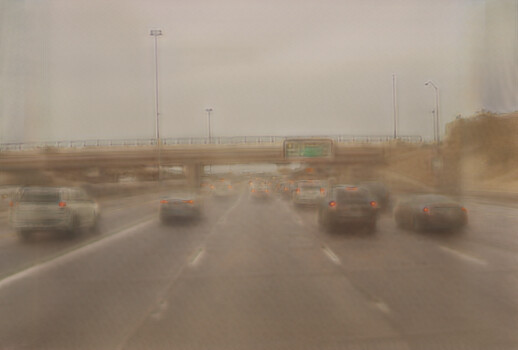} &
\includegraphics[width=\gridcellw]{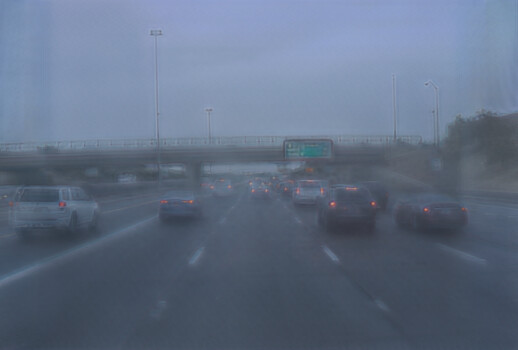} \\
\end{tabular}
\vspace{-2mm}
\caption{\textbf{Appearance transfer grid.}
Each row renders the same scene geometry under different appearance embeddings.
\emph{Column~1}: original render. \emph{Column~2}: base color ($\mathbf{a}{=}\mathbf{0}$).
\emph{Columns~3--7}: renders using the appearance embedding from the reference image shown in the top row.
The appearance embedding transfers the global color tone and lighting mood while preserving scene geometry.}
\label{fig:style_grid}
\end{figure*}

\subsection*{Additional Cross-Appearance Results}

We provide additional cross-appearance qualitative results on different Waymo segments in Figure~\ref{fig:qual_cross_supp1} and Figure~\ref{fig:qual_cross_supp2}.

\input{figures/qual_cross_supp1}

\input{figures/qual_cross_supp2}

\end{document}